\pdfoutput=1

\documentclass[11pt]{article}

\usepackage{naacl2021}

\usepackage{times}
\usepackage{latexsym}

\usepackage[T1]{fontenc}

\usepackage[utf8]{inputenc}

\usepackage{microtype}

\usepackage{latexsym}
\usepackage{bm}
\usepackage{subfigure}
\usepackage{multirow}
\usepackage{adjustbox}
\usepackage{array}
\usepackage{CJKutf8}
\usepackage{makecell}

\title{Addressing the Vulnerability of NMT in Input Perturbations}
\author{Weiwen Xu\raisebox{4pt}{\small $1,2$}~\thanks{~~Work was done when the author was a staff in Institute for Infocomm Research, A*STAR.}~, Ai Ti Aw\raisebox{4pt}{\small $1$}~\thanks{~~Corresponding Author}~, Yang Ding\raisebox{4pt}{\small $1$}, Kui Wu\raisebox{4pt}{\small $1$}, Shafiq Joty\raisebox{4pt}{\small $3$}\\
  \raisebox{4pt}{\small $1$}Institute for Infocomm Research, A*STAR\\
  \raisebox{4pt}{\small $2$}The Chinese University of Hong Kong\\
  \raisebox{4pt}{\small $3$}Nanyang Technological University\\
    {\tt wwxu@se.cuhk.edu.hk}\qquad \\{\tt \{aaiti, ding\_yang, wuk\}@i2r.a-star.edu.sg}\\  {\tt srjoty@ntu.edu.sg}
}

\begin{document}
\maketitle
\begin{abstract}
Neural Machine Translation (NMT) has achieved significant breakthrough in performance but is known to suffer vulnerability to input perturbations. As real input noise is difficult to predict during training, robustness is a big issue for system deployment. In this paper, we improve the robustness of NMT models by reducing the effect of noisy words through a Context-Enhanced Reconstruction (CER) approach. CER trains the model to resist noise in two steps: (1) perturbation step that breaks the naturalness of input sequence with made-up words; (2) reconstruction step that defends the noise propagation by generating better and more robust contextual representation. Experimental results on Chinese-English (ZH-EN) and French-English (FR-EN) translation tasks demonstrate robustness improvement on both news and social media text. Further fine-tuning experiments on social media text show our approach can converge at a higher position and provide a better adaptation.
\end{abstract}

\section{Introduction}
Recent techniques~\cite{bahdanau2014neural,wu2016google,NIPS2017_7181} in NMT have gained remarkable improvement in translation quality.  However, robust NMT that is immune to real input noise remains a big challenge for NMT researchers. Real input noises can exhibit in many forms such as spelling and grammatical errors, homophones replacement, Internet slang, new words or even a valid word used in an unfamiliar or a new context. Unlike humans who can easily comprehend and translate such texts, most NMT models are not robust to generate appropriate and meaningful translations in the presence of such noises, challenging the deployment of NMT system in real scenarios. 

\begin{table}
\begin{CJK*}{UTF8}{gbsn}
    \centering
    \small
    \begin{tabular}{p{7mm}<{\centering}|p{63mm}}
    \hline
        Input & 通宵打游戏上分贼快 \\ \hline
        Ref. & It’s super-fast to gain scores when playing games over the night. \\ \hline 
          MT& Play the game all night and take points thief fast. \\ \hline
        CER  & Play games all night to score points quickly. \\ \hline
    \end{tabular}
     \begin{tabular}{p{7mm}<{\centering}|p{63mm}}
    \hline
        Input & 我已剪短了我的发,剪断了惩罚,剪一地伤透我的尴尬。。。。 \\ \hline
        Ref. & I have cut my hair, i cut off the punishment, i away the awkwardness that hurt me. \\ \hline 
          MT & I got my punishment, got rid of my embarrassment. \\ \hline
        CER  & I cut short my hair , cut off punishment , and cut off my embarrassment  that hurts me. \\ \hline
    \end{tabular}
    \caption{Examples of NMT's vulnerability in translating text containing noisy words (``zei"  $\rightarrow$ ``thief", ``chengfa" $\rightarrow$ ``punishment"). CER mitigates the effect of noisy words.}
    \label{tab:noisy_example}
    \end{CJK*}
\end{table}

Noisy words have long been discussed in previous work. \citet{aw-etal-2006-phrase} proposed the normalization approach to reduce the noise before translation. \citet{tan-etal-2020-morphin, tan-etal-2020-mind} addressed the character-level noise directly in the NMT model. Though these approaches addressed the effect of noisy words to some extent, they are limited to spelling errors, inflectional variations, and other noises definable during training. In addition, strong external supervision like a parallel corpus of noisy text translation or dictionary containing the translation of those noisy words are hard and expensive to obtain; they are also not practical in handling real noises as noisy words can exhibit in random forms and cannot be fully anticipated during training. 

\citet{belinkov2018synthetic} pointed out NMT models are sensitive to small input perturbations and if this issue is not addressed, it will continue to bottleneck the translation quality. In such cases, not only the word embeddings of perturbations may cause irregularities with the local context, the contextual representation of other words may also get affected by such perturbations~\cite{liu-etal-2019-robust}. This phenomenon 
applies to valid words in unfamiliar context as well, which will also cause the translation to fail as illustrated in Table~\ref{tab:noisy_example} (case 2). 

In this paper, we define “noisy word” as a valid or invalid word that is uncommonly used in the context or not observed frequently enough in the training data. When encoding a sentence with such a noisy word, the contextual representation of other words in the sentence are affected by the ``less jointly trained" noisy word embeddings. We refer this process as ``noise propagation". Noise propagation can extend to the decoder and finally distort the overall translation. 

The main intuition of our proposed method is to minimize this noise propagation and reduce the irregularities in contextual representation due to these words via a \textit{Context-Enhanced Reconstruction} (CER) approach. To reduce the sensitivity of contextual  towards noisy words in the encoder, we inject made-up words randomly to the source side of the training data to break the text naturalness. We then use a \textit{Noise Adaptation Layer} (NAL) to enable a more stable contextual representation by minimizing the reconstruction loss. In the decoder, we add perturbations with a semantic constraint and apply the same reconstruction loss. Unlike adversarial examples which are crafted to cause the target model to fail, our perturbation process does not have such constraint and does not rely on a target model. Our input perturbations are randomly generated, representing any types of noises that can be observed in real-world usage. This makes the perturbation process generic, easy and fast.
Following~\cite{cheng-etal-2018-towards}, we generate semantically related perturbations in the decoder to increase the diversity of the translations.

Together with NAL, our model shows its ability to resist noises in the input and produce more robust translations. Results on ZH-EN and FR-EN translation significantly improve over the baseline by +1.24 (MT03) and +1.4 (N15) BLEU on news domain, and +1.63 (\textit{Social}), +1.3 (\textit{mtnt18}) on social media domain respectively. Further fine-tuning experiments on FR-EN social media text even witness an average improvement of +1.25 BLEU over the best approach. 

\section{Related Work}
\paragraph{Robust Training:}Robust training has shown to be effective to improve the robustness of the models in computer vision~\cite{szegedy2013intriguing}. In Natural Language Processing, it involves augmenting the
training data with carefully crafted noisy examples: semantically equivalent word substitutions~\cite{alzantot2018generating}, paraphrasing~\cite{iyyer-etal-2018-adversarial,ribeiro-etal-2018-semantically}, character-level noise~\cite{ebrahimi-etal-2018-hotflip,tan-etal-2020-morphin,tan-etal-2020-mind}, or perturbations at embedding space~\cite{miyato2016adversarial,DBLP:conf/mm/LiangLCYSC20}. Inspired by~\citet{ijcai2017-562} that nicely captures the semantic interactions in discourse relation, we regard noise as a disruptor to break semantic interactions and propose our CER approach to mitigate this phenomenon. We make up “noisy” words randomly to act as random noise in the input to break the text naturalness. Our experiment demonstrates its superiority in multiple dimensions.

\paragraph{Robust Neural Machine Translation:} Methods have been proposed to make NMT models resilient not only to adequacy errors~\cite{lei-etal-2019-revisit} but also to both natural and synthetic noise. Incorporating monolingual data into NMT has the capacity to improve the robustness~\cite{sennrich-etal-2016-improving,edunov-etal-2018-understanding,cheng-etal-2016-semi}. Some non data-driven approaches that specifically designed to address the robustness problem of NMT~\cite{sperber2017toward, ebrahimi-etal-2018-adversarial, wang-etal-2018-switchout, karpukhin2019training, cheng-etal-2019-robust, cheng-etal-2020-advaug} explored effective ways to synthesize adversarial examples into the training data. \citet{belinkov2018synthetic} showed a structure-invariant word representation capable of addressing multiple typo noise. \citet{cheng-etal-2018-towards} used adversarial stability training strategy to make NMT resilient to arbitrary noise. \citet{liu-etal-2019-robust} added an additional phonetic embedding to overcome homophone noise.

Meanwhile, \citet{michel-neubig-2018-mtnt} released a dataset for evaluating NMT on social media text. This dataset was used as a benchmark for WMT 19 Robustness shared task~\cite{li-EtAl:2019:WMT1} to improve the robustness of NMT models on noisy text. We show our approach also benefits the fine-tuning process using additional social media data. 

\section{Approaches}

\begin{figure*}[ht]
    \small
    \centering
     \includegraphics[scale=0.4]{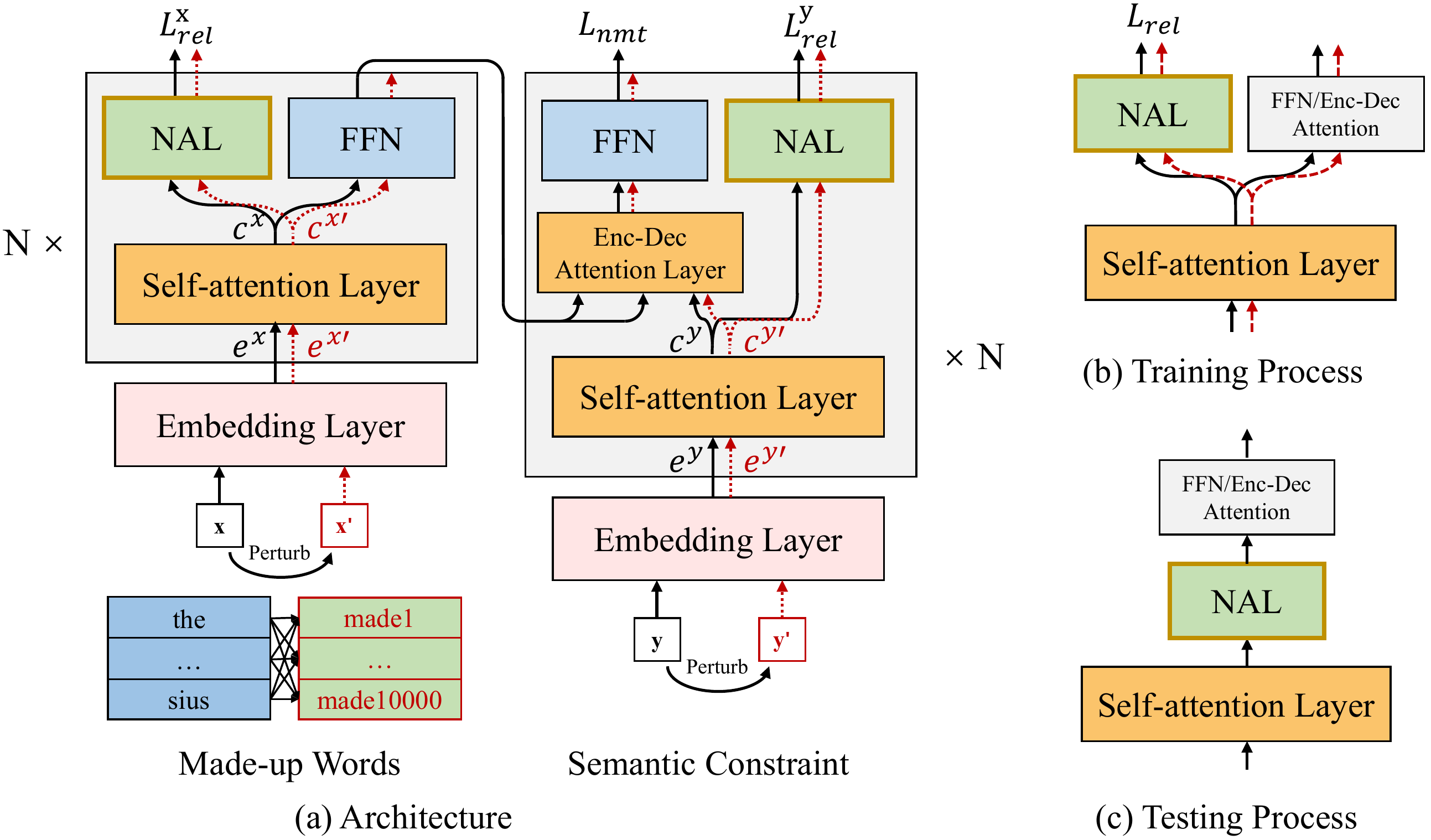} 
       \caption{ The architecture of CER (a), and the use of NAL in training (b) and testing (c). The solid lines indicate the flow for original input, while the dotted lines for noisy input, generated in the perturbation step.}
    \label{fig:model}
\end{figure*}

We propose a Context-Enhanced Reconstruction (CER) approach to learn robust contextual representation in the presence of noisy words through a perturbation step and a reconstruction step in both encoder and decoder during model training. Figure~\ref{fig:model} shows the architecture.

The perturbation step automatically inserts made-up words in the input sequence $\mathbf{x}$ to generate a noisy example $\mathbf{x'}$. The noisy example mimics input where text naturalness is broken due to the noisy words. Similarly, we perturb the output sequence $\mathbf{y}$ to $\mathbf{y'}$ using a semantic constraint to generate noisy examples for the decoder to have more diversity in the translations.

The reconstruction step in the model aims to restore the contextual representation $\mathbf{c^{x'}}$ of $\mathbf{x'}$ to be similar to its corresponding original contextual representation $\mathbf{c^x}$ in the encoder. Specifically, under the Transformer architecture (Figure~\ref{fig:model}), the reconstruction step aims to stabilize and minimize the disruption of attention distribution for a word over the whole input in the presence of inserted noise. The stabilization is needed for both clean and noisy words as both of their contextual representations are affected. For a noisy word, reconstruction reduces the attention to itself and encourages the construction of the contextual representation to leverage more on its clean neighbors. For clean words, reconstruction works as a denoise module to mitigate the interference of noisy words. For $\mathbf{c^{y'}}$ in the decoder, the aim is to generate more examples with similar context as  $\mathbf{c^y}$. The reconstruction helps to normalize the contextual representation of semantically similar words.

\subsection{Perturbing Input Text with Noise}
\label{section_breaking}
We insert made-up words, representing any kinds of noise, to disturb the contextual representation during training. To create those words, we build a made-up dictionary $\mathcal{D}^-_x$ with $M$ made-up words. As shown in Figure~\ref{fig:model}(a), made-up words are simply indexed slots in $\mathcal{D}^-_x$, whose embeddings are randomly initialized with no prior restriction and updated during training just as valid words. During the perturbation step, we randomly select multiple positions in each input sequence based on probability $\sigma_x$ and replace the words with any arbitrary made-up words in $\mathcal{D}^-_x$.

For the decoder, as the aim is not to insert noise but to increase the diversity of translation, we add small perturbations with a semantic constraint to make the model robust. Specifically, we randomly select multiple positions in each target sequence with a probability $\sigma_y$ and perturb the corresponding words. For the word $y_i$ chosen to be perturbed, we create a dynamic set $\mathcal{V}_{y_i}$ consisting of $m$ words having the highest cosine similarity with it (excluding $y_i$). We average the embeddings of the words in $\mathcal{V}_{y_i}$ as the perturbation for $y_i$.
\begin{equation}
\small
    \mathcal{V}_{y_i} = \mathop{top\_m}\limits_{y_j \in \mathcal{D}_y, j \ne i} (cos(e^{y_i}, e^{y_j}))
\end{equation}
\begin{equation}
\small
    e^{y_i'} = \frac{1}{m}\sum_{ y_j \in  \mathcal{V}_{y_i}}e^{y_j}
\end{equation}
Where $\mathcal{D}_y$ is the target dictionary, $e^{y_j}$ is the target word embedding for $y_j$ and $e^{y_i'}$ is the perturbed embedding for $y_i$.

\subsection{Reconstructing Contextual Representation}
\label{section_reconstruction}
As the injected noise in $\mathbf{x'}$ affects the self-attention mechanism in producing correct contextual representation, we regularize the contextual representation using a Noise Adaptation Layer (NAL) immediately after the self-attention layer as depicted in Figure~\ref{fig:model}(a). This NAL is trained together with the NMT model and used as a reconstruction module during testing (See Figure~\ref{fig:model}(b),(c)). 

Formally, let $\mathbf{c}^\mathbf{x}_l$ and $\mathbf{c}^\mathbf{x'}_l$ be the outputs of the self-attention in the $l$-th encoder layer for $\mathbf{x}$ and $\mathbf{x'}$ respectively. We train the NAL by:
\begin{equation}
\label{loss_e}
\small
\mathcal{L}^x_{nal}(\bm{\theta}^x_{nal}) = \frac{1}{|S|}\sum_{(\mathbf{x},\mathbf{y})\in S} \sum_{l=1}^N ||\mathbf{c}^\mathbf{x}_l - {\rm NAL}(\mathbf{c}^\mathbf{x'}_l) ||^2
\end{equation}
Where $\bm{\theta}^x_{nal}$ are parameters of NAL, $S$ is the training corpus and $N$ is the encoder layer size. Given $\mathbf{c^{x'}}$, NAL attempts to output a more correct contextual representation guided by $\mathbf{c^{x}}$. We use a single layer feed-forward network (FFN) in \cite{NIPS2017_7181} as our NAL implementation. Similarly, the reconstruction loss for decoder is:
\begin{equation}
\label{loss_d}
\small
\mathcal{L}^y_{nal}(\bm{\theta}^y_{nal}) = \frac{1}{|S|}\sum_{(\mathbf{x},\mathbf{y})\in S} \sum_{l=1}^N ||\mathbf{c}^\mathbf{y}_l  - {\rm NAL}(\mathbf{c}^\mathbf{y'}_l ) ||^2
\end{equation}

\subsection{Model Training}
\label{section_training}
We apply the perturbation step at the embedding layer, see Figure~\ref{fig:model}. The inserted noise in $\mathbf{x'}$ and $\mathbf{y'}$ would also receive gradient from the final loss function and update just like other clean words. NAL is added at each Transformer layer where the outputs are only used to calculate the reconstruction loss and not passed to the next layer. On the other hand, the output of FFN is propagated to the next layer as usual. The reconstruction step mainly serves as a stabilizer to prevent the noise from propagating. 

The final training objective $\mathcal{L}$ is the combination of the above three loss functions,  the original translation loss, the reconstruction loss for the encoder and the reconstruction loss for the decoder. Both $\lambda_x$ and $\lambda_y$ are set empirically to count for the relative importance.
\begin{equation}
\label{loss}
\small
\mathcal{L} = \mathcal{L}_{nmt}(\bm{\theta}_{nmt}) + \lambda_x \mathcal{L}^x_{nal}(\bm{\theta}^x_{nal}) + \lambda_y \mathcal{L}^y_{nal}(\bm{\theta}^y_{nal})
\end{equation}

\section{Experiment Settings}
Experiments are conducted on ZH-EN and FR-EN translation tasks for both news and social media domains. We also use social media text to fine-tune the NMT systems on FR-EN.
\subsection{Data}
\noindent \textbf{ZH-EN:} The training data consists of 1.25M sentence pairs extracted from LDC. For news domain, we use NIST MT02 as the development set and select the best model to test MT03, MT04, MT05, MT06 and MT08 news test sets. For social media domain, we create a test set (\emph{Social}) consisting of 2000 sentences with three human annotated references. The source sentences are collected from public social media platforms in four Chinese-speaking regions: Mainland China, Hong Kong, Taiwan and Singapore~\footnote{Available at \url{https://github.com/wwxu21/CER-MT.}}.

\noindent \textbf{FR-EN:} We use the same datasets as \citet{michel-neubig-2018-mtnt}. The training set consists of 2.16M sentence pairs extracted from  \textit{europarl-v7} and \textit{news-commentary-v10}.  We use the \textit{newsdiscussdev2015} as development set and evaluate the model on two news test sets, \textit{newstest2014} (N14) and \textit{newsdiscusstest2015} (N15). We also evaluate on two social media test sets: \textit{mtnt18}~\cite{michel-neubig-2018-mtnt} and \textit{mtnt19}~\cite{li-EtAl:2019:WMT1}.

\noindent \textbf{FR-EN Fine-Tuning:} We use the noisy training set (\textit{mtnttrain}) provided by \citet{michel-neubig-2018-mtnt} to fine-tune the FR-EN model.

We use fairseq's implementation of Transformer~\cite{ott2019fairseq}. In evaluation, we report case-insensitive tokenized BLEU for ZH-EN~\cite{papineni2002bleu} and {\tt sacre-BLEU}~\cite{post2018call} for FR-EN. Following \citet{michel-neubig-2018-mtnt}, we do not use development set but only report best results on three social media test sets. 

We segment the Chinese words using THULAC~\cite{li-sun-2009-punctuation} and tokenize both French and English words using {\tt tokenize.perl}\footnote{https://github.com/moses-smt/mosesdecoder}. We apply BPE~\cite{sennrich-etal-2016-neural} to get sub-word vocabularies for the encoder and decoder, both with 20K merge operations.

The hyper-parameters setting is the same as {\tt transformer-base} in \cite{NIPS2017_7181} except that we set dropout rate as 0.4 in all our experiments. Our proposed models are trained on top of Transformer baseline for efficiency purpose, where additional parameters from the embeddings of $\mathcal{D}^-_x$ and ReL are uniformly initialized. The madeup dictionary size $M$ is set to 10,000. The size of  dynamic set $m$ is set to 3. The probability $\sigma_x$ and $\sigma_y$ are both set to 0.1 and balance coefficient $\lambda_x$ and $\lambda_y$ are both set to 1.

\begin{table*}[]
    \centering
    \small
    \begin{tabular}{l||c|cccccc|c}
    \hline
    Model     & MT02 (DEV) & MT03 &  MT04 &  MT05 &  MT06 &  MT08 &News Ave. & \textit{Social}\\ \hline
    \multicolumn{8}{c}{ \textit{Existing systems}} \\\hline
      \citet{wang-etal-2018-switchout}   & 47.13& 46.68 &47.41  &46.66 &46.62 & 38.46 &45.17 &23.20  \\
      \citet{cheng-etal-2018-towards}   &46.10 &44.07 & 45.61 &43.45 &44.44 &34.94 &42.50 &21.27\\
      \citet{cheng-etal-2019-robust}   &47.06 &46.48 & 47.39 &46.58 &46.95 &37.38 &44.96 &22.74 \\
      \hline   
      \multicolumn{8}{c}{ \textit{Our systems}} \\ \hline 
            Transformer   &46.98 &46.35 & 47.27 &46.35 &46.77 &38.20 & 45.00 &22.41 \\ 
    + CER-Enc & 47.65 & 46.72 & 47.53 & 47.06 & 47.04 & 38.53 &45.38 &23.81  \\
      + CER & \textbf{48.34} & \textbf{47.59} & \textbf{48.21} & \textbf{47.29} & \textbf{47.64} & \textbf{39.33} & \textbf{46.01}& \textbf{24.04}  \\
      \hline
    \end{tabular}
    \caption{Case-insensitive BLEU scores (\%) on ZH-EN translation. MT02 is our development set.}
    \label{tab:zh-en_translation}
\end{table*}

\subsection{Baseline Models}
\label{sec:baseline}
We use Transformer as our baseline. 

\noindent \textbf{ZH-EN}: We compare with \citet{wang-etal-2018-switchout, cheng-etal-2018-towards, cheng-etal-2019-robust}. \citet{wang-etal-2018-switchout} use a data augmentation approach by randomly replacing words in source and target sentences with other in-dictionary words. \citet{cheng-etal-2018-towards} use adversarial stability training to make NMT resilient to noise. \citet{cheng-etal-2019-robust} employ a white-box approach to synthesize adversarial examples. 

\noindent \textbf{FR-EN}: In addition to \citet{wang-etal-2018-switchout}, we compare with \citet{michel-neubig-2018-mtnt, zhou-EtAl:2019:WMT, vaibhav-etal-2019-improving} on FR-EN or FR-EN Fine-Tuning tasks. \citet{michel-neubig-2018-mtnt} do the first benchmark of the noisy text translation tasks in three languages. \citet{vaibhav-etal-2019-improving} leverage effective synthetic noise to make NMT resilient to noisy text. We implement their approach on Transformer backbone. For a fair comparison, we limit the data to train back-translation models only with \textit{mtnttrain}. \citet{zhou-EtAl:2019:WMT} adopt a multitask transformer architecture with two decoders, where the first decoder learns to denoise and the second decoder learns to translate from the denoised text. They adopt the approach proposed by \citet{vaibhav-etal-2019-improving} to synthesize the noisy text for their first decoder.

We do not compare our model with~\cite{berard-calapodescu-roux:2019:WMT,helcl-libovick-popel:2019:WMT} as they use much more out-domain data, a great number of monolingual data and a bigger Transformer model, and hence not comparable with our experimental settings.

\begin{table}[]
    \centering
    \small
    \begin{tabular}{l||cc|cc}
    \hline
    Model     & N14 & N15 &  \textit{mtnt18} & \textit{mtnt19}  \\ \hline 
    \multicolumn{4}{c}{ \textit{Exising systems}} \\ \hline
      
      \citeauthor{wang-etal-2018-switchout}   &29.2 &31.1 &25.0 &28.1  \\
      \citeauthor{michel-neubig-2018-mtnt}   &28.9 &30.8 &23.3 &26.2 \\
      \citeauthor{zhou-EtAl:2019:WMT}*  & N.A. & N.A. & 24.5 & \textbf{30.3}\\
      \hline   
      \multicolumn{4}{c}{ \textit{Our systems}} \\ \hline 
      Transformer  &29.7 &31.0 &25.2 &28.0  \\ 
      + CER-Enc   &30.4 &31.7  & 26.1 & 28.7 \\
      + CER   &\textbf{30.7} &\textbf{32.4}  &\textbf{26.5} & 29.1 \\
      \hline
    \end{tabular}
    \caption{{\tt sacreBLEU} (\%) on FR-EN translation task. *\citeauthor{zhou-EtAl:2019:WMT} use more data to train their model.}
    \label{tab:fr-en_translation}
\end{table}

\section{Results and Analysis}
\subsection{Comparison with Baseline Models}

Table~\ref{tab:zh-en_translation} and Table~\ref{tab:fr-en_translation} show the performance on ZH-EN and FR-EN tasks. We show the results of applying CER only to the encoder (+ CER-Enc), and to both the encoder and decoder (+ CER). 

 As illustrated, our approach improves the news text translations on all test sets for both ZH-EN and FR-EN and outperforms the Transformer baseline in terms of average BLEU by +1.01 and +1.2 on ZH-EN and FR-EN respectively, illustrating the superiority of our approach. 

The performance on social media test sets shows significant improvement with up to +1.63 BLEU over Transformer and +0.84 BLEU over the best approach~\cite{wang-etal-2018-switchout} on ZH-EN. For FR-EN, our model outperforms~\citet{wang-etal-2018-switchout} by +1.5 and +1.0 BLEU on \textit{mtnt18} and \textit{mtnt19} respectively. \citet{zhou-EtAl:2019:WMT} use \textit{mtnttrain} and TED~\cite{qi-etal-2018-pre:} to synthesize noisy sentences for their first decoder, hence effectively they are exploiting in-domain data during training and thus not quite a fair comparison in the evaluation. Nevertheless, CER still significantly outperforms~\citet{zhou-EtAl:2019:WMT} by +2.0 BLEU on \textit{mtnt18}.

\begin{figure}
    \centering
    \includegraphics[scale=0.6]{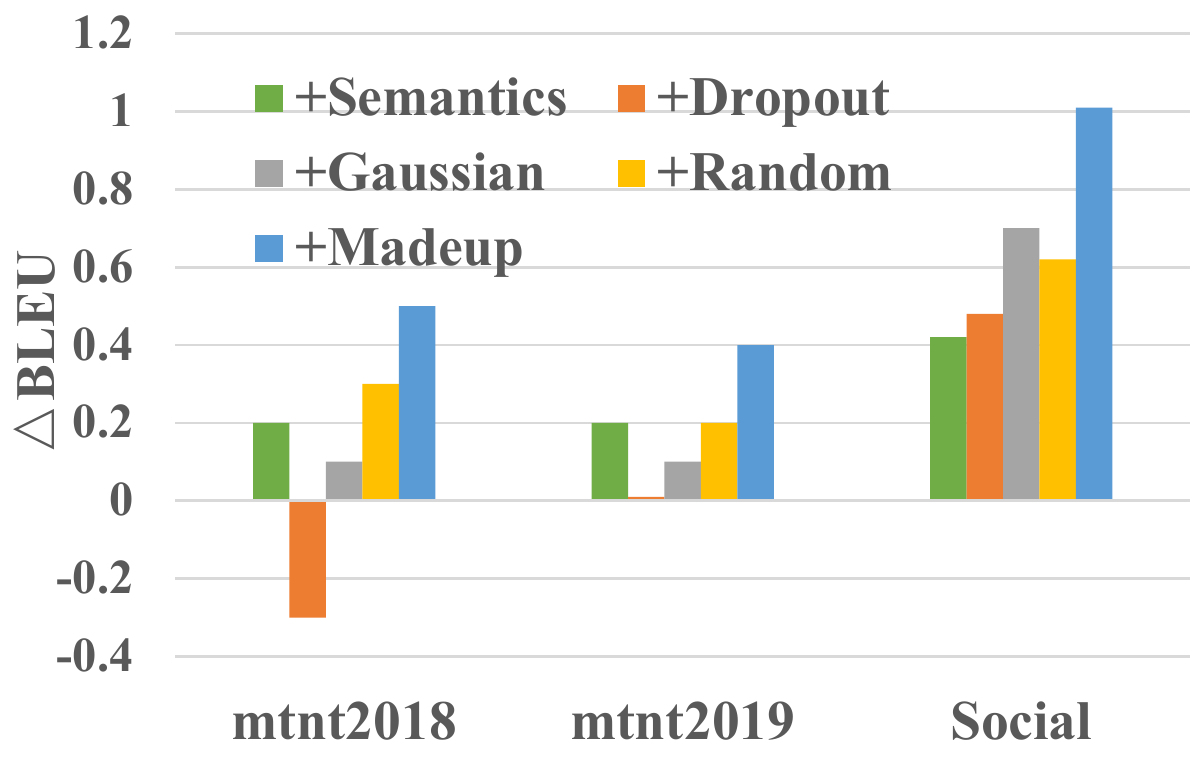}
    \caption{BLEU improvements compared to Transformer baseline shown in Table~\ref{tab:zh-en_translation} and Table~\ref{tab:fr-en_translation} when applying noise-insertion methods.}
    \label{fig:noise}
\end{figure}

\subsection{Effect of Noise}
We investigate the effect of different noise-insertion methods by dynamically inserting noise into the source side of the original training set using different strategies with a same probability $\sigma_x$. 

\noindent \textit{Madeup}: Our approach to add made-up words.

\noindent \textit{Semantics}: We test our semantic constraint in the decoder to assess if it benefits the encoder.  

\noindent \textit{Dropout}: We replace word embeddings with all-0 vectors, similar to enlarging the dropout rate.

\noindent \textit{Gaussian}: Following the feature-level perturbations of \citet{cheng-etal-2018-towards}, we add the Gaussian noise to a word embedding to simulate the noise. 

\noindent \textit{Random}: We replace a word with an arbitrary word in the dictionary. This would result in a valid word being placed in an unreasonable context.

Figure~\ref{fig:noise} shows the BLEU improvement of various noise-insertion methods on social media test sets. We find that nearly all kinds of noise-insertion methods improve the robustness of MT with the exception of  \textit{Dropout}. Since we have already set the dropout rate to an optimal rate, inserting additional \textit{Dropout} noise does not increase but decreases the performance. As shown, \textit{Madeup} improves the performance nearly twice than the rest of the noise-insertion methods. We conjecture \textit{Semantics}, \textit{Dropout} and \textit{Gaussian} may be small and not diverse enough to simulate the real noisy words. Both \textit{Random} and \textit{Madeup} can break the text coherence. However, \textit{Random} uses a random in-dictionary word, which can place a valid word in an unreasonable context and cause its embedding to update in a wrong direction. In fact, this method improves the robustness of NMT models at the cost of those replaced words. Our \textit{Madeup} can entirely avoid this cost as we use made-up words to work as noisy words and does not cause any context change of all in-dictionary words.

\begin{figure}
    \centering
    \includegraphics[scale=0.55]{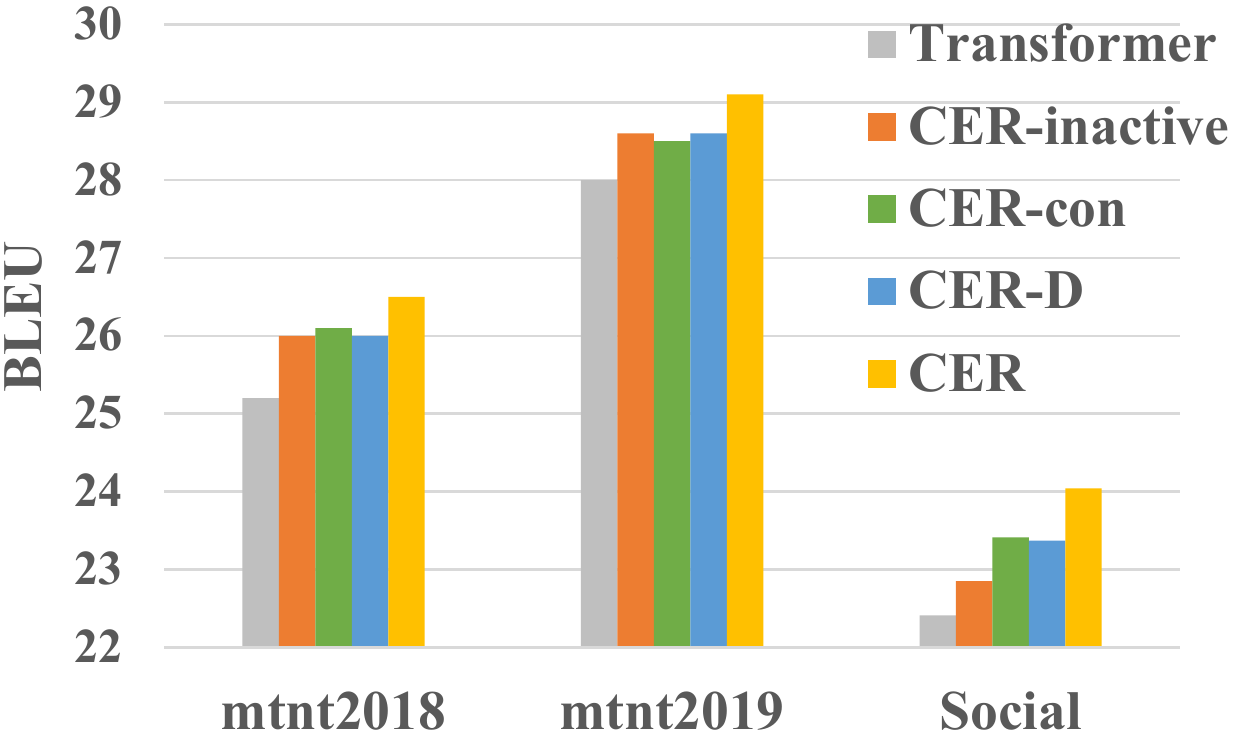}
    \caption{BLEU scores of CER variants.}
    \label{fig:variants}
\end{figure}

\begin{table}
    \centering
    \small
    \begin{tabular}{l||cc}
    \hline
     Model     &  mtnt18 &   mtnt19 \\ \hline
     \multicolumn{3}{c}{ \textit{Existing systems}} \\ \hline
      
        \citeauthor{michel-neubig-2018-mtnt}   &30.3 & N/A \\
        \citeauthor{wang-etal-2018-switchout} & 35.1 & 36.7\\
        \citeauthor{zhou-EtAl:2019:WMT} & 31.7 & 32.8\\
        \citeauthor{vaibhav-etal-2019-improving} & 36.0 & 37.5\\
        \hline
        \multicolumn{3}{c}{ \textit{Our systems}} \\ \hline
         Transformer (Base)  & 25.2 & 28.1 \\
       +FT & 35.2 & 37.4 \\
    +FT w/ CER& \textbf{37.3} & \textbf{38.7} \\ \hline
    \end{tabular}
    \caption{{\tt sacreBLEU} on FR-EN fine-tuning task.}
    \label{tab:fr-en_FT}
\end{table}
\subsection{Effect of NAL}
To further gain insights on how NAL helps improve the robustness of NMT models. We create three variants to aid our analysis:

\noindent \textbf{CER-inactive}: We do not activate NAL at testing time. The contextual representation is feed directly into later FFN. This variant is to test the effectiveness of NAL.

\noindent \textbf{CER-con}: We remove NAL but only add a constraint to ensure $\{\mathbf{c^x},\mathbf{c^{x'}}\}$ and $\{\mathbf{c^y},\mathbf{c^{y'}}\}$ to be close respectively at training time. This forces the self-attention layer to reconstruct the correct contextual representation itself. This variant is to demonstrate the necessity to set apart the context generation module (self-attention layer) and the reconstruction module (NAL).

\noindent \textbf{CER-D}: We borrow the adversarial stability training strategy proposed in \citet{cheng-etal-2018-towards} here. In this variant, NAL is replaced by a discriminator and $\bm{\theta}_{nal}^x$ and $\bm{\theta}_{nal}^y$ are changed to the adversarial learning loss in \citet{cheng-etal-2018-towards}. The purpose is to assess the effectiveness of NAL and the discriminator in context reconstruction.

Figure~\ref{fig:variants} shows the results of the three variants on three social media test sets. From the figure, we make the following observations.

\textit{NAL is effective at Test Time.} The activation of NAL at test time helps to produce more reliable contextual representation. Notably, NAL gains +1.19 BLEU on \textit{Social}.

\textit{NAL needs to be learnt separately.} As shown in CER-con, by forcing self-attention layer to do both tasks (context generation and reconstruction), the performance improvement gets affected by at least 0.4 BLEU.

\textit{NAL is more effective than a discriminator to guide reconstruction.} The improvements are less significant in all test sets when using a discriminator (CER-D) comparing to CER. Therefore, we can conclude that NAL is more effective than a discriminator to reconstruct the perturbed contextual representation and CER outperforms all variants.

\subsection{FR-EN Fine-Tuning on Social Media Text}
 We fine-tune the same Transformer model in Table~\ref{tab:fr-en_translation} with the social media data \textit{mtnttrain} (\emph{+FT}) and further include CER in the fine-tuning (+FT w/ CER). Table~\ref{tab:fr-en_FT} shows our performance (\emph{+FT w/ CER}) with other four fine-tuning approaches on \textit{mtnttrain}. It shows that our CER also benefits the fine-tuning process and outperforms all the approaches in two noisy test sets. Specifically, it gains +2.1 and +1.3 BLEU over \emph{+FT} on \textit{mtnt18} and \textit{mtnt19} and outperforms~\citet{vaibhav-etal-2019-improving} by +1.3 and +1.2 BLEU respectively. 
 
\subsection{Experiments on Large-Scale Datasets}
We first train a ZH-EN baseline model using 25M sentence pairs, which are mainly in news domain. Similar to the setting in Table~\ref{tab:fr-en_FT}, we apply both simple finetuning (+FT) and our CER (+ FT w/ CER) approach using 125K social media training data. We evaluate those models on \textit{Social}. We also include the performance of Google Translate~\footnote{https://translate.google.com/} here to show the competitiveness of our baseline model.

As shown in Table~\ref{tab:large-scale}, our CER approach can still benefit the fine-tuning process even on the strong baseline. It should be noted that the baseline has already maintained high robustness with large-scale training data  where improvement in such a model is hard to obtain. In fact, 125K in-domain data can only contribute to 1.55 BLEU improvement. Under this circumstance, the 0.26 BLEU improvement brought by CER should be highly valued considered no additional fine-tuning data is used.

\begin{table}
    \centering
    \large
    \begin{tabular}{l|l||c}
    \hline
      \multicolumn{2}{l||}{Model}     &  \textit{Social} \\ \hline
      
        \multicolumn{2}{l||}{Google Translate}    &38.59  \\
        \hline \hline
        \multirow{3}{*}{Ours} &Baseline  & 39.01  \\
      & ~~ +FT & 40.56 (+3.97\%) \\
    & ~~ +FT w/ CER& \textbf{40.82 (+4.64\%)}  \\ \hline
    \end{tabular}
    \caption{Case-insensitive BLEU scores (relative improvement) on large-scale ZH-EN translation system.}
    \label{tab:large-scale}
\end{table}

\section{Conclusions}
In this work, we propose an approach to reduce the vulnerability of NMT models to input perturbations. Our input perturbation is easy, fast and not specific to a target victim model. Experimental results show our proposed approach improves the robustness on both news and social media text and helped to improve the translation of real input.

\section{Acknowledgments}
The work was supported in part by Defence Science and Technology Agency (DSTA), Singapore.
\bibliography{naacl2021}
\bibliographystyle{acl_natbib}

\end{document}